# Research Project: What cat is that? A re-id model for feral cats

**Master of Artificial Intelligence and Machine Learning**

Victor Caquilpan

Supervised by Prof. Ian Reid[1] and Stefan Podgorski[1]

**Abstract**

*Feral cats exert a substantial and detrimental impact on Australian wildlife, placing them among the most dangerous invasive species worldwide. Therefore, closely monitoring these cats is essential labour in minimizing their effects. In this context, the potential application of Re-Identification (re-ID) emerges to enhance monitoring activities for these animals, utilizing images captured by camera traps. This project explores different CV approaches to create a re-ID model able to identify individual feral cats in the wild. The main approach consists of modifying a part-pose guided network (PPGNet) model, initially utilized in the re-ID of Amur tigers, to be applicable for feral cats. This adaptation, resulting in PPGNet-Cat, which incorporates specific modifications to suit the characteristics of feral cats' images. Additionally, various experiments were conducted, particularly exploring contrastive learning approaches such as ArcFace loss. The main results indicate that PPGNet-Cat excels in identifying feral cats, achieving high performance with a mean Average Precision (mAP) of 0.86 and a rank-1 accuracy of 0.95. These outcomes establish PPGNet-Cat as a competitive model within the realm of re-ID.*

## 1. Introduction

Feral cats are recognized as one of the most hazardous invasive mammals, being one of the primary predators of birds and other small mammals (Kosicki 2021, Stokeld 2016). Due to its devastating effect in the population of native animals, the monitoring and management of these animals becomes a relevant need, which requires accurate knowledge of their population density and their individual behaviour (Rees et al. 2019).

Feral cats are mostly cryptic, nocturnal, and trap-shy, which increases the constraints to perform accurate population estimates (Legge et al. 2017). In this context, the utilization of camera traps emerges as a key method for monitoring and estimating population sizes. However, despite the abundance of captured images, individually identifying cats remains a laborious and intricate task (Sparkles et al. 2021). Mostly, their identification is done manually based on unique pelage markings, which is a time-consuming activity and, in some cases, due to the poor quality of images and the lack of specific differentiable patterns (e. g. black cats) the identification is difficult to carry on (Rees et al. 2019).

Re-ID refers to the task of recognizing and matching individuals or objects across multiple images or cameras. The main goal behind re-ID is to identify the same individual as it appears or moves in various locations and under different conditions, making possible their searching and tracking (Wei et al. 2022). The field of person re-ID has gained substantial popularity in recent years, driven by its diverse applications in management and security, as noted by Zheng et al. (2016) and Wei et al. (2022). However, the application of these models to animals has not been equally explored in the same manner. This gap may be attributed to the significant challenges associated with animal re-ID, including the lack of proper labelled data, and the more substantial intra-identity variations and smaller inter-identity distances in the case of several animals. Compounding these challenges is the fact that animals may be subject to various unconstrained imaging conditions, stemming from variations in body pose, view angle, illumination, and occlusion making more difficult their identification (Zhang et al. 2021, Zheng et al. 2022).

Developing an effective re-ID model capable of accurately identifying feral cats under diverse conditions would contribute to the management of feral cat populations and addressing an extensive range of questions in the study of ecosystem function, community, and population dynamics (Schneider et al. 2018). In that way, it might be possible to enhance population density estimations, recognizing individual-level patterns, and understanding the displacement and distribution of cats across various locations. In this project, different experiments are conducted to develop a suitable re-ID model to identify feral cats in the wild, being a part-pose based model the main architecture explored.

## 2. Literature review

In the realm of re-ID advancements, various methodologies have been investigated, encompassing metric learning, part-based models, GAN-based models, transformers, and video-based models, among others, each offering distinct advantages and drawbacks (Wenyu et al. 2022, Zheng et al. 2022). Most of these approaches have been implemented in person re-ID, with some adaptations to animal contexts (Liu, Zhang and Guo 2019, Tuia et al. 2022, Zuerl et al. 2023). In this field, part-based models have garnered significant attention due to their ability to robustly extract features as well as keeping these features over different poses, being important in animal re-ID contexts.

There is a scarcity of research specifically addressing re-ID of felines, with only a few available projects on the subject. In the study conducted by Yang et al. (2021), they implemented a modified version of YOLOv5 to identify feral cats using images from camera traps in Australia, achieving a mAP of 0.77. On the other hand, in the Computer Vision for Wild Conservation Challenge (CVWC2019), Liu, Zhang, and Guo (2019) presented the Part-Pose Guided (PPGNet) model to identify Amur tigers in the ATRW dataset. This model outperformed the state-of-the-art and finished top-1 in the re-ID competitions at CVWC2019, obtaining a mAP of 0.72 under cross-camera conditions. Due to its outstanding performance and the resemblance between tigers and cats, this latter method was selected as the primary approach for this project.

---

1. School of Computer Science, The University of Adelaide

## 3. Methodology

To develop this project, multiple experiments were conducted to perform re-ID models. In this section, the dataset and the most important approaches are described.

**Data**

Obtaining adequate re-ID data for animal identification poses a significant challenge, primarily due to the difficulty of acquiring labelled images. In 2019, the CVWC2019 introduced the Amur Tiger Re-ID in the Wild (ATRW) dataset, which emerged as one of the most comprehensive re-ID datasets for animals at that time, comprising 3,649 images featuring 92 distinct entities (Li et al. 2020). In addition, these images bring a set of 15 key-points of different body parts of tigers (Figure 1), which might help in a pose estimation process.

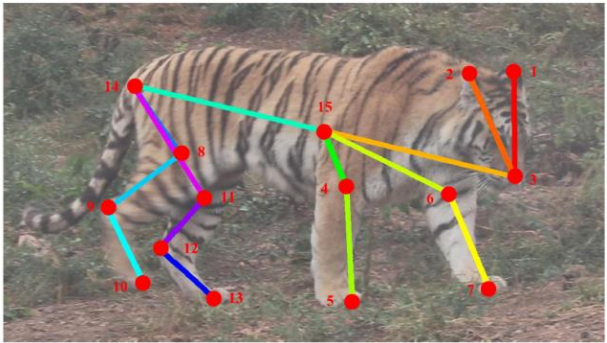

**Figure 1**. One image sample of the ATRW dataset (Li et al. 2020).

This dataset was utilised as a baseline to explore different re-ID approaches. It was chosen due the proficient quality of the images, as well as the similarities in the morphology of tigers with feral cats, serving as the foundational reference point for developing our re-ID feral cat model.

In the specific context of feral cats, our primary experiments were conducted using a dataset derived from monitoring activities in Western Australia (referred to as the WA-feral dataset). This dataset comprises 3,120 images capturing 10 distinct individual feral cats, collected from various camera traps and locations.

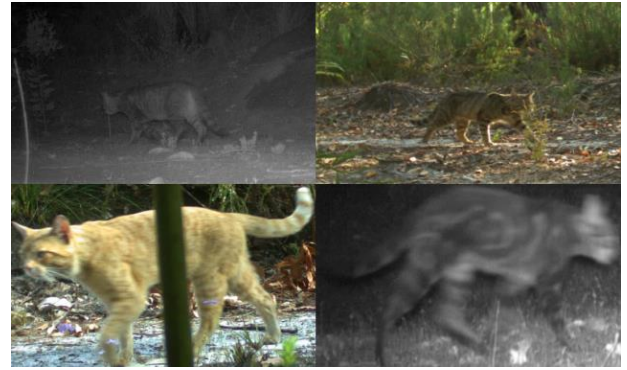

**Figure 2.** Collection sample of feral cat images. The image located in the top-left quadrant depicts a nocturnal scene, while the top-right corresponds to a daytime setting. The bottom-left features an image with occlusion, and the bottom-right quadrant showcases a blurred image.

These images are captured directly by camera traps, featuring resolutions of Full HD and 2K, being capable of capturing images both during the day and at night (Figure 2). The key attributes are outlined below:

- Most feral cat images are captured during the night, as these animals exhibit nocturnal behaviour (Legge et al. 2017).
- In several instances, the cat's body appears small in the images, particularly when the cats are at a distance from the camera.
- In nocturnal images, the deficiency of illumination adversely impacts image quality, posing challenges in their identification.
- Blurred images are prevalent, a consequence of the continuous movement of the cats.
- Occasionally, objects such as branches or sticks obstruct the cat's body, leading to instances of occlusion.

Unfortunately, as multiple of these images given in the feral cat dataset correspond to incomplete body (e. g. just the tail or the head of the cat), it is decided to remove these cases. Additionally, adhering to the methodology outlined by Yang et al. (2021), since numerous images are associated with the same capture sequence, we encounter multiple closely resembling images. However, to mitigate the risk of overfitting, a subset of these images were excluded. In that way, the WA-feral dataset was reduced to 752 useful images.

Furthermore, to evaluate the adaptability of our models in a distinct scenario, an extra dataset featuring images of feral cats was acquired from Rees et al. (2023). This dataset (called Victoria-feral dataset) comprises images captured during monitoring activities in National parks of Victoria, Australia, from where we could obtain a total of 21 cats across 35 entities. It served as a supplementary source to evaluate the adaptability of our top model.

**Preprocess data**

Based on the nature of feral cat images a set of processes were undertaken to modify the input to take advantage of their main properties. Below, there is a general description of these steps:

**Crop images**: Since most of these images are not centred to the cat body, it becomes necessary to crop these images to match the specific dimensions of each cat. To streamline this task, the Yolov8 detection module was integrated to automate the cropping process.

**Splitting entities**: Li et al. (2020) note that tigers display unique stripe patterns on both their left and right sides. Due to the infrequency of encountering tigers showcasing both sides simultaneously in the environment, the researchers regard each side of the tigers as distinct entities (classes). Since the similarities in the morphology of tigers with feral cats, it was decided to follow this approach. In addition, as night time images are lower resolution and contain more noise (due to the lower light levels and higher gain of the sensor) than daytime images, classical re-ID models may not be adapted to these nocturnal conditions (Zhang, Yuan and Wang, 2019). Considering these facts, it was decided to split entities based on the cat's orientation (left and right) as well as the timestamp of the image (day and night). In the Figure 3, we can see an example of this splitting.

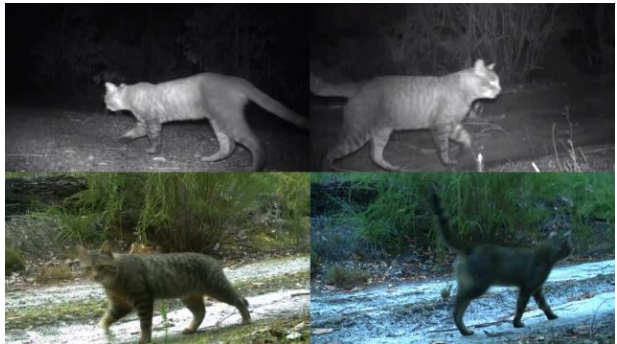

**Figure 3**. Different entities for the same cat. Top quadrants are images of the same cat in nocturnal condition for left and right side, respectively. Bottom images are daytime images for the same individual in different sides.

**Splitting train/test**: To deploy our re-ID, the data was split into a train and test subset in a ratio of 60/40. Table 1 contains the characteristics of each subset.

**Table 1**. Characteristics of feral cat dataset.

|  | Train | Test |
|---|---|---|
| N° Cats | 6 | 4 |
| N° Entities | 20 | 10 |
| N° Images | 424 | 328 |

It is important to observe that the entities present in the test subset are not included in the training subset, aligning with the structure of the ATRW dataset (Li et al. 2020).

**Inclusion of key-points:** Based on the work of Liu, Zhang and Guo (2019), they deal pose variations of tigers based on key-points (body joints). This strategy was employed to minimize the adverse effects of pose variance. To explore the use of part-posed models, 15 key-points were manually added to the images where it was possible to recognize them, trying to replicate the ATRW dataset conditions (Figure 1). The definition of each one of these key-points is explained in Li et al. (2020).

**Data augmentation**: Given the diverse characteristics of the images employed in this project, various data augmentation techniques are applied to simulate a range of different image conditions. Some of these techniques include:

- Implementation of Gaussian blur filter: Useful for simulating the effect of low light or specific atmospheric conditions.
- Noise injection: Addition of noise in the image by a Gaussian distribution.
- Random erasing: Implementation of random dropout in the input data space to combat challenges related to occlusion (Shorten and Khoshgoftaar 2019).
- Perspective: Modification of the perspective of images to add some distortion modifications that might occur due to different camera angles (Wang et al. 2020).
- Rotate: Rotation of images of cats to consider different cats' orientation (Shorten and Khoshgoftaar 2019).

**Model architecture**

The main model employed in this work is PPGNet. Figure 4 shows its architecture. The key aspects of this approach are outlined in the following points:

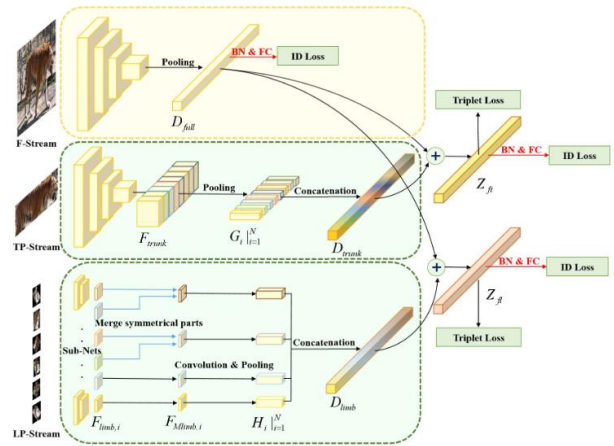

**Figure 4.** PPGNet model architecture (Liu, Zhang and Guo, 2019).

- The model comprises a full stream (F-Stream) and two partial streams (TP-Stream and LP-Stream). During the training phase, the partial streams serve as regulators to guide the full stream in learning both global and local features. However, during the inference step, only the F-Stream is employed, resulting in a more lightweight and faster prediction process.
- The F-Stream utilizes the complete image of the tiger, whereas the TP-Stream and LP-Stream utilize images of the trunk and limbs, respectively. To extract the sections corresponding to the trunk and limbs, rectangular shapes were generated using the key-points provided in the ATRW dataset to encompass the corresponding body segments.
- The backbone used to extract the main features is ResNet152 in the F-Stream and ResNet34 for the partial streams.
- The F-Stream produces the $D_{full}$ embedding, meanwhile from the partial streams, $D_{trunk}$ and $D_{limbs}$ are obtained, which are summed to $D_{full}$ to create $Z_{ft}$ and $Z_{fl}$, respectively.

- $D_{full}$, $Z_{ft}$ and $Z_{fl}$ are passed to a combination of cross-entropy (ID loss) and triplet loss.

These are some of the features presented in this model. More details can be found in Liu, Zhang and Guo (2019). Based on these properties, PPGNet-Cat was proposed as an adapted version of PPGNet specifically tailored for the context of feral cats.

### PPGNet-Cat

In this project, PPGNet was applied to the task of identifying feral cats, with additional modifications made to enhance its baseline performance specifically for this context, resulting in the creation of PPGNet-Cat. These modifications are listed below.

- To crop the limbs, rectangles are utilized, where the width is obtained as a ratio of the height. This relation was adapted to the cat's body to enclose the width of the limbs, which are thinner than tigers. The relation height/width is 1/3.
- To crop the trunk, an axis-aligned rectangle is implemented, assuming that tigers are mostly in a horizontal position. However, we observed that in multiple cases cats are not properly in a horizontal position, therefore the rectangle was rotated to be aligned with the orientation of the cat's body (Figure 5).

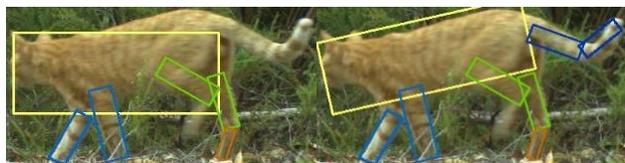

**Figure 5**. Body parts detection. Left image employs the approach of Liu, Zhang and Guo (2019) to crop the trunk and limbs. Right image considers the modification of trunk (rotated) and inclusion of tail.

- The use of limbs proposed in PPGNet comes from the availability of key-points in the ATRW dataset. Despite the cropped limbs and trunks cover most of the body of the cat, the tail is not represented in this process, even considering that some specific patterns may be obtained from this body part. For that reason, the inclusion of two extra key-points to get the proximal and distal tail was considered (Figure 5). To extract these body parts, the same approach to extract the limbs was used.
- Since in some images it is not possible to get the proper image of the trunk and limbs, PPGNet includes black images as input in these cases. However, as a pretrained version of ResNet is used, the output embedding ($D_{trunk}$ and $D_{limb}$) for these black images corresponds to noise (biases provided mostly by the Batch normalization layers). Here, an output of zeros is defined in $D_{trunk}$ and $D_{limb}$ when black images are entered to avoid this noise.

The main architecture of PPGNet-Cat is represented in the Appendix 1. Complementary, multiple approaches were addressed mostly in the field of contrastive learning, being ArcFace loss the principal complementary approach used in this project. ArcFace is a loss function commonly used in face recognition tasks, which enhances the discriminative power of the learned features by imposing a margin in an angular space of the feature embeddings (Deng et al. 2019). Due to the effectiveness in handling face recognition challenges (intra-class variations and inter-class similarities), it was decided to perform some experiments using this approach.

### Main experiments

Experiments were undertaken to address the key research questions in this study, using PPGNet, ArcFace and PPGNet-Cat as the main approaches, testing their performance over the WA-feral dataset. Also, there is included the implementation of a simple model based on ResNet152 as a comparison baseline. Additionally, a series of experiments were undertaken to assess the following points:

- Measure the improvement provided by PPGNet-Cat in comparison to PPGNet.
- Evaluate the impact of treating day and night images as separate entities on accurate identification.
- Also, assess the significance of categorizing cat images based on their sides (left/right) in terms of their contribution.
- Asses the main challenging cases of our best model.

Pytorch was the deep learning framework used to deploy the different experiments in this research. In addition, the models were trained over a NVIDIA GeForce RTX 2080i card. The experiments involved running tests over a wide spectrum of epoch numbers, ranging mostly from 100 to 300.

### Evaluation

Re-ID models can be evaluated in different ways, being some of the most common metrics mAP and Rank-1. Unlike Rank-1, mAP measures the average retrieval performance with multiple ground truths. In this way, for mAP the first ground truth matters as well as the subsequent matches, which could be more difficult to identify properly (Ye et al. 2021). For that reason, the main metric used in this study is mAP with Rank-1 as a complementary metric.

### 4. Results

The presented results showcase the optimal outcomes derived from this experimentation process. Table 2 summarises the most relevant results obtained in the test subset of feral cats. From that, we can observe that PPGNet-Cat gives us the best performance in terms of mAP and Rank-1, outperforming the other tested approaches.

**Table 2**. Main results using different implementations of ArcFace, PPGNet and PPGNet-Cat.

| Model | mAP | Rank-1 |
|---|---|---|
| ResNet152 | 0.44 | 0.62 |
| ResNet152 + ArcFace | 0.60 | 0.83 |
| PPGNet* | 0.77 | 0.91 |
| PPGNet* + ArcFace | 0.82 | 0.95 |
| PPGNet-Cat | **0.86** | **0.95** |
| PPGNet-Cat + ArcFace | 0.82 | 0.94 |

*PPGNet is an adapted version of the model presented by Liu, Zhang and Guo, (2019).

In a general perspective, ArcFace shows its utility in refining accurate cat identification. However, upon integrating ArcFace into PPGNet-Cat, the results indicate that this combination did not improve the model's performance. Certain experiments incorporating ArcFace into PPGNet-Cat suggest that using a limited number of epochs (e.g., less than 100) yields improved predictions, converging faster to a better solution (however, not the optimal). Nonetheless, as the model trains for more epochs, ArcFace becomes ineffective in facilitating identification (however, it may be possible to find better combination of parameters to match ArcFace implementation).

Table 3 illustrates several key modifications made over PPGNet and their impact on the performance of our ultimate PPGNet-Cat model. This demonstrates how several of these adjustments had a positive impact on the overall performance of the model.

**Table 3**. Main adaptations over PPGNet-Cat.

| Model Adaptation | mAP | Rank-1 |
|---|---|---|
| Base PPGNet | ~0.75 | ~0.89 |
| + Random erasing | ~0.82 | ~0.93 |
| + Forcing zeros in black images | ~0.825 | ~0.95 |
| + Rotated trunk | ~0.83 | ~0.95 |
| + Inclusion of tail (PPGNet-Cat) | **~0.86** | **~0.95** |

To depict our results, we employed the Uniform Manifold Approximation and Projection (UMAP) technique to reduce the dimensionality of the embeddings obtained from the test subset images, diminishing them from 2,560 to 2 dimensions. UMAP is a dimension reduction technique, which has proven effective in preserving a greater extent of the global structure within the input data while concurrently ensuring superior computational efficiency (McInnes and Healy, 2018). This visualization is possible to see in Figure 6.

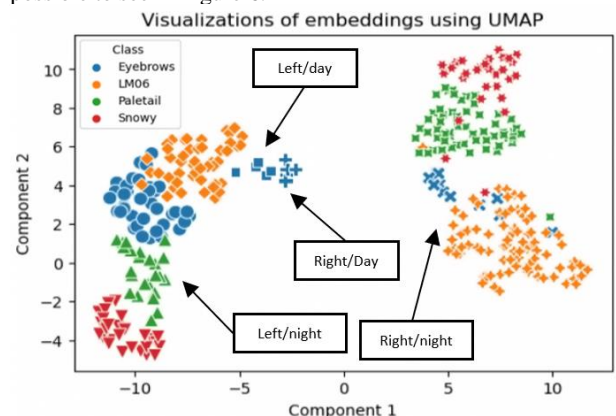

**Figure 6**. Representation of embeddings using UMAP.

Figure 6 shows the allocation of 328 images (test subset). Each colour in the figure represents a distinct cat, while each marker shape corresponds to a specific entity. Notably, markers of the same colour and shape are predominantly clustered together, signifying accurate predictions, however, there are instances where points deviate from their primary cluster, indicating cases where the model faced challenges in proper identification. The visualization reveals two primary groups of points that correspond to entities on different sides (left and right) in nocturnal images, observing there is a considerable distinction between them. In addition, there is a small cluster of points in the centre, corresponding to day images. The small number of points is due there are just 10-day time images in the subset test. A reduced distance is noted between these two groups, suggesting that our model may not perform effectively on these images. This could be attributed to the limited number of daytime images available during the training process.

In Figure 7, utilizing a randomly selected image from the test subset (depicting cat "Eyebrows") as a query image, the ranking-7 is presented, showcasing the seven closest images to the query. The initial six matches align with images of the same class, whereas the last image is a mismatch, belonging to a different class. This discrepancy may be attributed to the pronounced similarity in stripe patterns between "Eyebrows" and "LM06," resulting in a failure case.

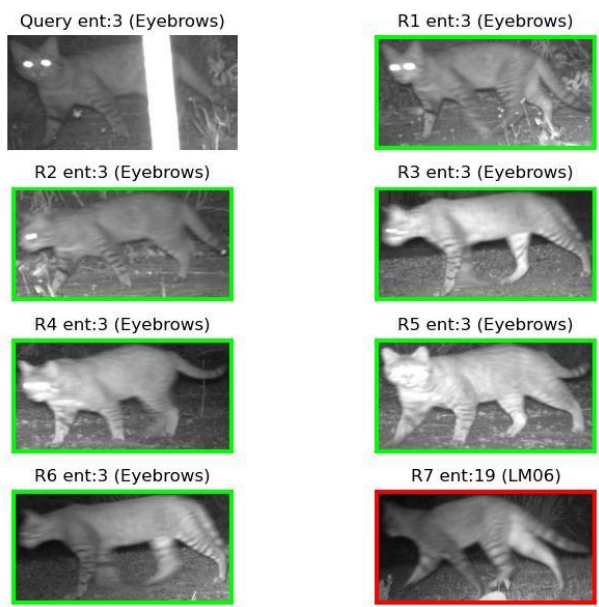

**Figure 7**. Example of a ranking-7 for a query image.

In addition to the aforementioned outcomes, to show the impact of segregating entities based on their side and time conditions, PPGNet-Cat underwent testing across various partition settings. The outcomes of this experimentation are presented in Table 4. Consistently, the identical model was employed in each case, undergoing training under each distinct partition configuration. Notably, the results highlight that employing side and time partitions yields optimal outcomes in terms of mAP and Rank-1, justifying its use.

**Table 4.** Performance of PPGNet-Cat over different partition setting.

| Setting | N° entities | mAP | Rank-1 |
|---|---|---|---|
| Without partition | 4 | 0.73 | 0.93 |
| Day/Night partition | 5 | 0.78 | 0.91 |
| Left/Right partition | 8 | 0.48 | 0.81 |
| Left/Right Day/Night partition | 10 | **0.86** | **0.95** |

Additionally, to evaluate the adaptability of PPGNet-Cat to other contexts, it was performed an inference process over the Victoria-feral dataset. The application of PPGNet-Cat over this dataset resulted in mAP of 0.73 and a Rank-1 accuracy of 0.87. Although the performance metrics are comparatively lower than those obtained with the WA-feral dataset, it may be justified by the increased number of individuals present in this new dataset.

Ultimately, it is worth noting that despite PPGNet-Cat being a relatively sizable model with 197 million parameters (~789 megabytes), the inference model itself is more lightweight, with 70 million parameters (~283 megabytes). This characteristic renders it more efficient for execution during the prediction step. An updated version of this model is available in this GitHub repository. Some demos are available to demonstrate the application of PPGNet-Cat.

## 5. Discussion

Based on the previous results, we can determine that PPGNet-Cat can be employed in the identification of feral cats, obtaining promising results in terms of mAP and Rank-1. It has been demonstrated that the use of a part-posed model can be utilised for this purpose having proper annotated data. A notable drawback of this kind of approaches lies in their reliance on accurately annotated key-points. Typically, another algorithm, such as pose estimation, is employed for this task, however, there exists a risk of introducing additional noise, potentially resulting in reduced model generalization (Wei et al. 2022). In contrast, PPGNet effectively addresses this issue by exclusively utilizing key-points presented just in the training datasets, allowing for a more comprehensive focus and mitigating the risk of noise interference.

Analysing the instances of misidentification in our predictions allows us to pinpoint the primary challenges of PPGNet-Cat model. Figure 8 illustrates two examples of misidentifications. In the first case (top images), the model erroneously identifies two tabby cats with similar patterns on their bodies as the same individual. This may be based on the high similarity on their patterns. In the second scenario, two similar-looking cats are considered, but both images include the presence of a white stick (a baiting equipment located in the front of some camera traps), potentially causing the model to recognize the stick patterns as indicative of the same individual. This hypothesis is reinforced by the absence of such images in the training dataset.

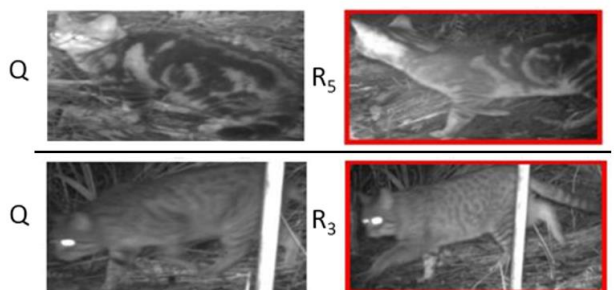

**Figure 8.** Display two instances of mismatches, where Q denotes the query image, and $R_3$ and $R_5$ represent rankings 3 and 5, respectively.

In addition to the aforementioned scenarios, PPGNet-Cat, while striving to identify both the global and local features of each entity through visual patterns, encounters challenges in instances that include blurry images or cats with intricate patterns that are difficult to discern (such as complete black cats). It is crucial to emphasize that the WA-feral dataset is relatively small, consisting of only 10 cats. Providing a larger dataset for training could enhance the model's ability to identify cats in a broader range of contexts.

On the other hand, as illustrated in Table 4, employing four partitions has the potential to enhance the identification capability of our model. Nevertheless, dividing individuals into four distinct entities may introduce challenges in accurately counting the actual number of individuals during monitoring. Despite these limitations, the model can still provide valuable insights to improve the identification process. Another limitation is related to this study concentrates on the visual patterns of cats presented in their left and right sides. However, images depicting the front or back of the cats are not conducive to their accurate identification.

As a future work, incorporating a larger dataset for training our model is under consideration. Due to constraints in time and data availability, preparing a more robust input dataset has not been feasible thus far. A substantial dataset has the potential to address key challenges highlighted in this research. It's worth noting that several hyperparameters were adopted from Liu, Zhang, and Guo (2019), specifically tailored to the ATRW context. However, adjusting and fine-tuning some of these parameters could potentially enhance prediction accuracy.

## 6. Conclusion

During this research, the development of a re-ID model for feral cats in the wild have been developed. This model called PPGNet-Cat is a part pose based model, which has demonstrated notable success in identifying wild cats, showcasing a commendable mAP of 0.86 and a Rank-1 accuracy of 0.95, outperforming other tested approaches.

Analysing the outcomes across both the WA-feral and Victoria-feral datasets, it is evident that PPGNet-Cat demonstrates effective generalization in varied contexts. However, it is essential to acknowledge instances where the model encounters challenges in achieving accurate identifications. Exploring the inclusion of a more diverse dataset in our training data might present an avenue for improved performance, particularly in addressing key challenges faced by the model.

Despite their limitations, their promising results signify the potential utility of PPGNet-Cat in the monitoring of feral cats in Australia. The model shows potential in aiding essential tasks associated with cat management and tracking, offering valuable insights for wildlife conservation initiatives, and enhancing the efficiency of the re-identification process, which is currently predominantly carried out manually. The robust performance metrics emphasize its effectiveness as a tool for improving the identification and monitoring processes within feral cat populations, both in Australia and in other regions where monitoring these populations is a pertinent responsibility.

## 7. Acknowledgements

The realization of this project was made possible through the collaboration and support of Thylation and the Department of Biodiversity, Conservation, and Attractions (DBCA) of Western Australia. These entities played a crucial role by offering primary guidance on the state of feral cat management and providing the necessary images for the development of the model.

## 8. References

Deng, J, Guo, J, Xue, N & Zafeiriou S 2019, 'ArcFace: Additive Angular Margin Loss for Deep Face Recognition', in *2019 IEEE/CVF Conference on Computer Vision and Pattern Recognition (CVPR)*, Long Beach, CA, USA, pp. 4685-4694.

Kosicki, JZ 2021, 'The impact of feral domestic cats on native bird populations. Predictive modelling approach on a country scale', *Ecological Complexity*, vol. 48, pp. 1-10.

Legge, S, Murphy, BP, McGregor, H, Woinarski, JCZ, Augusteyn, J, Ballard, G, Baseler, M, Buckmaster, T, Dickman, CR, Doherty, T, Edwards, G, Eyre, T, Fancourt, BA, Ferguson, D, Forsyth, DM, Geary, WL, Gentle, M, Gillespie, G, Greenwood, L & Hohnen, R 2017, 'Enumerating a continental-scale threat: How many feral cats are in Australia?', *Biological Conservation,* vol. 206, pp. 293–303.

Li, S, Li, J, Tang, H, Qian, R, Lin, W 2020, 'ATRW: A Benchmark for Amur Tiger Re-identification in the Wild', in *MM '20: Proceedings of the 28th ACM International Conference on Multimedia*, New York, USA.

Liu, C, Zhang, R & Guo, L 2019, 'Part-Pose Guided Amur Tiger Re-Identification', in *2019 IEEE/CVF International Conference on Computer Vision Workshop (ICCVW)*, IEEE, Seoul, Korea (South).

McInnes, L & Healy, J 2018, 'UMAP: Uniform Manifold Approximation and Projection for Dimension Reduction', *ArXiv e-prints*, pp. 1-63.

Rees, MW, Pascoe, JH, Wintle, BA, Le Pla, M, Birnbaum, EK, & Hradsky, BA 2019, 'Unexpectedly high densities of feral cats in a rugged temperate forest', *Biological Conservation*, vol. 239, no. 108287, DOI:10.1016/j.biocon.2019.108287.

Schneider, S, Taylor, GW, Linquist, S, Kremer, SC 2019, 'Past, present and future approaches using computer vision for animal re-identification from camera trap data', *Methods in Ecology and Evolution*, vol. 10, no. 4 pp. 461–470.

Shorten, C & Khoshgoftaar TM 2019, 'A survey of Image Data Augmentation for Deep Learning', *Journal of Big Data*, vol. 6, no. 60, pp. 1-48.

Sparkes, J, Fleming, PJS, McSorley, A, & Mitchell, B 2021, 'How many feral cats can be individually identified from camera trap images? Population monitoring, ecological utility and camera trap settings', *Ecological Management & Restoration*, vol. 22, no. 3, pp. 246-255, DOI:10.1111/emr.12506.

Stokeld, D, Frank, A, Hill, B, Choy, J, Mahney, T, Stevens, A, Young, S, Rangers, D, Rangers, W & Gillespie, G 2016, 'Multiple cameras required to reliably detect feral cats in northern Australian tropical savanna: an evaluation of sampling design when using camera traps', *Wildlife Research*, vol. 42, pp. 642-649.

Tuia, D, Kellenberger, B, Beery, S, Costelloe, BR, Zuffi, S, Risse, B, Mathis, A, Mathis, MW, van Langevelde, F, Burghardt, T, Kays, R, Klinck, H, Wikelski, M, Couzin, ID, van Horn, G, Crofoot, MC, Stewart, CV & Berger-Wolf, T 2022, 'Perspectives in machine learning for wildlife conservation', *Nature Communications*, vol. 13, no. 792.

Wang, K, Fang, B, Qian, J, Yang, S, Zhou, X & Zhou, J 2020, 'Perspective Transformation Data Augmentation for Object Detection', *IEEE Access*, vol. 8, pp. 4935-4943.

Wei, W, Wenzhong, Y, Enguang, Z, Yunyun, Q & Lihua, W 2022, 'Person re-identification based on deep learning — An overview', *Journal of Visual Communication and Image Representation*, vol. 82, no. 103418, DOI:10.1016/j.jvcir.2021.103418.

Yang, Z, Sinnott, R, Ke, Q & Bailey, J 2021, 'Individual Feral Cat Identification through Deep Learning', in *DBCAT'21: 2021 IEEE/ACM 8th International Conference on Big Data Computing, Applications and Technologies*, pp. 101-110.

Ye, M, Shen, J, Lin, G, Xiang, T, Shao, L & Hoi, S 2021, 'Deep Learning for Person Re-Identification: A Survey and Outlook', *IEEE Transactions on Pattern Analysis and Machine Intelligence*, vol. 44, no. 6, pp. 2872-2893.

Zhang, J, Yuan, Y & Wang, Q 2019, 'Night Person Re-Identification and a Benchmark', in *IEEE Access*, vol. 7, pp. 95496-95504, 2019, DOI:10.1109/ACCESS.2019.2929854.

Zhang, T, Zhao, Q, Da, C, Zhou, L, Li, L & Jiancuo, S 2021, 'YakReID-103: A Benchmark for Yak Re-Identification', *2021 IEEE International Joint Conference on Biometrics (IJCB)*, Shenzhen, China, 2021, pp. 1-8.

Zheng, L, Yang, Y, Hauptmann AG 2016, 'Person Re-identification: Past, Present and Future', *ArXiv (CoRR)*.

Zheng, Z, Zhao, Y, Li, A, Yu, Q 2022, 'Wild Terrestrial Animal Re-Identification Based on an Improved Locally Aware Transformer with a Cross-Attention Mechanism', *Animals*, vol. 12, no. 3503, pp. 1-14.

Zuerl, M, Dirauf, R, Koeferl, F, Steinlein, N, Sueskind, J, Zanca, D, Brehm, I, Fersen, L von & Eskofier, B 2023, 'PolarBearVidID: A Video-Based Re-Identification Benchmark Dataset for Polar Bears', *Animals*, vol. 13, no. 5, pp. 1-15.

# 9. Appendix

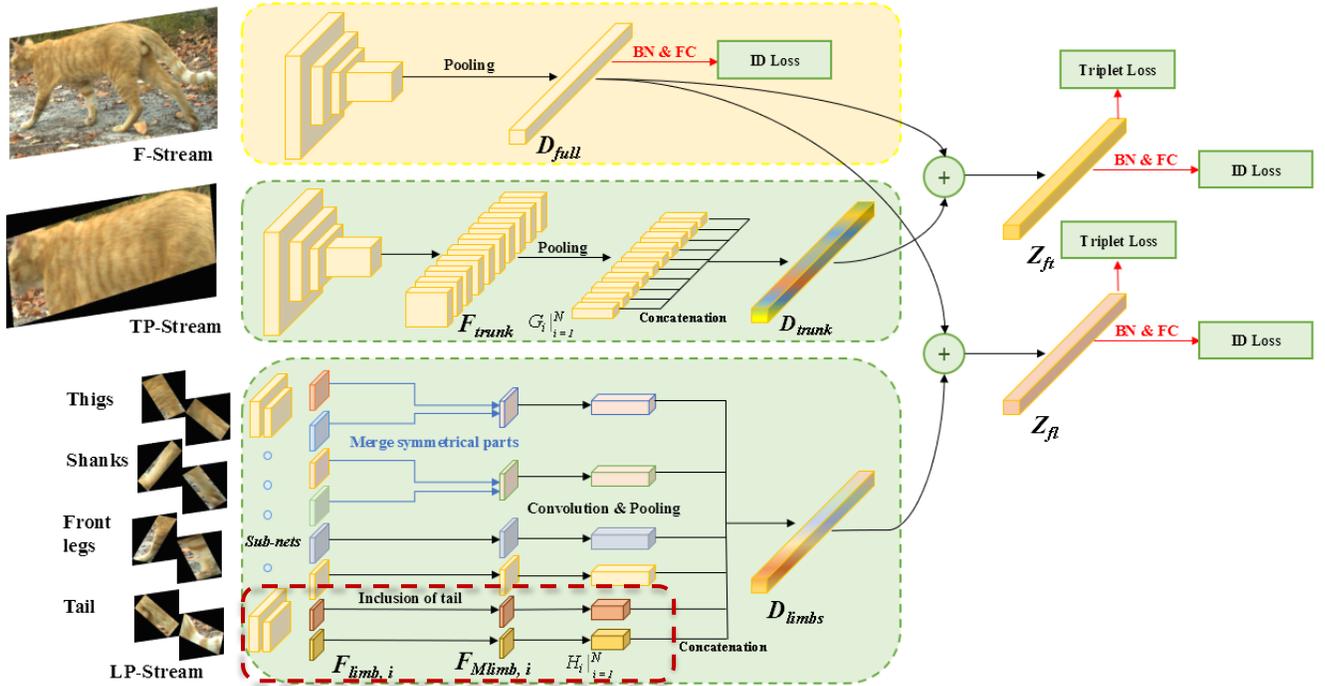

**Appendix 1**. PPGNet-Cat model architecture. Besides the use of limbs, in the LP-Stream two sub-networks were added to extract the feature maps corresponding to the proximal and distal tail) (red bounded box). The result of this extraction is reflected in two embeddings of 256 size each one, which is half the size of the embeddings for the limbs, maintaining the fundamental nature of the model. Finally, to preserve the core architecture of PPGNet, the embedding sizes throughout the entire model are increased from 2,048 to 2,560. This adjustment facilitates the integration of tail features into the final embedding, avoiding major modifications to the rest of the model.